\pdfoutput=1

\documentclass[11pt]{article}

\usepackage{naacl2021}
\usepackage{flushend}

\usepackage{times}
\usepackage{latexsym}

\usepackage{xcolor}
\usepackage{soul}

\usepackage[T1]{fontenc}

\usepackage[utf8]{inputenc}

\usepackage{microtype}
\usepackage{graphicx}

\title{Accountable Error Characterization}

\author{Amita Misra, Zhe Liu and Jalal Mahmud\\
  IBM-Research, Almaden \\
  San Jose, CA, USA \\
  {\tt amita.misra1|liuzh|jumahmud@ibm.com} \\}

\begin{document}
\maketitle
\begin{abstract}
Customers of machine learning systems demand accountability from the companies employing these algorithms for various prediction tasks. Accountability requires understanding of system limit and condition of erroneous predictions, as customers are often interested in understanding the incorrect predictions, and model developers are absorbed in finding methods that can be used to get incremental improvements to an existing system. Therefore, we propose an accountable error characterization method, AEC, to understand when and where errors occur within the existing black-box models. AEC, as constructed with human-understandable linguistic features, allows the model developers to automatically identify the main sources of errors for a given classification system. It can also be used to sample for the set of most informative input points for a next round of training. We perform error detection for a  sentiment analysis task using AEC as a case study. Our results on the sample sentiment task show that AEC is able to characterize erroneous predictions into human understandable categories and also achieves promising results on selecting erroneous samples when compared with the uncertainty-based sampling.
\end{abstract}

\section{Introduction}
\label{intro}

As machine learning is becoming the method of choice for many analytics functionalities in industry, it becomes crucial to be able to understand the limits and risks of the existing models. In favour of more accurate AI, the availability of computational resources is coupled with increasing dataset sizes that has resulted in more complex models. Complex models suffer from lack of transparency, which leads to low trust as well as the inability to fix or improve the models output easily.
 Deep learning algorithms are among the highly accurate and complex models. Most users of deep learning models often treat them as a black box because of its incomprehensible functions and unclear working mechanism \cite{LiuYW19}. However, customers’ retention requires  accountability for these systems \cite{Galitsky18a}. Interpreting and understanding what the model has learned, as well as the limits and the risks of the existing model have therefore become a key ingredient of a robust validation \cite{MontavonSM18}.

One line of research on model accountability examines the information learned by the model itself to probe the linguistic aspects of language learnt by the models \cite{shi-etal-2016,AdiKBLG17,GiulianelliHMHZ18,Belinkov19,LiuYW19}.
Other line of research gives machine learning models the ability to explain or to present their behaviours in understandable terms to humans \cite{DoshiVelez2017} to make the predictions more transparent, and trustworthy. However, very few studies set the focus on error characterization as well as automatic error detection and mitigation.
To address the above-mentioned gaps in characterizing model limits and risks, we seek to improve a model’s behavior by categorizing incorrect predictions using explainable linguistic features. To accomplish that, we propose a framework called Accountable Error Characterization (AEC) to explain the predictions of a neural network model by constructing an explainable error classifier. The most similar work to ours is by \cite{NushiKH18}. They build  interpretable decision-tree classifiers for summarizing failure conditions using human and machine generated features. In contrast, our approach builds upon  incorrect predictions on a separate set to provide insights into  model failure.

The AEC framework has three key components: A base neural network model,  an error characterization model, and a set of interpretable features that serve as the input to the error characterization model. The features used in the error characterization model are based on explainable linguistic and lexical features such as dependency relations, and various lexicons that have been inspired by prior art, which allows the users and model developers to identify when a model fails. The error characterization model also offer rankings of informative features to provide insight into where and why the model fails. 

By adding the error classification step on top of the base model, AEC can also be adopted to identify the highly confident error cases as the most informative samples for the next round of training. Although uncertainty based sampling can also be adopted to get the most informative samples {\cite{Lewis95,Cawley11,ShaoWL19}}, as it selects the examples with the least confidence, \newcite{GhaiLZBM20} show that uncertainty sampling led to an increasing challenge for annotators to provide correct labels. AEC avoids such problem by learning from error cases from a validation set. Our results show that AEC outperforms the uncertainty based sampling in terms of selecting erroneous predictions on a sample sentiment dataset (see Table~\ref{eval}).

We first present the overall framework of AEC to construct the error classifier, followed by the experiments and result. Finally, we conclude the paper with future directions and work in progress.

\section{Explainable Framework }
\label{method-sec}
\begin{figure}
\centering
	\includegraphics[height=5.0cm]{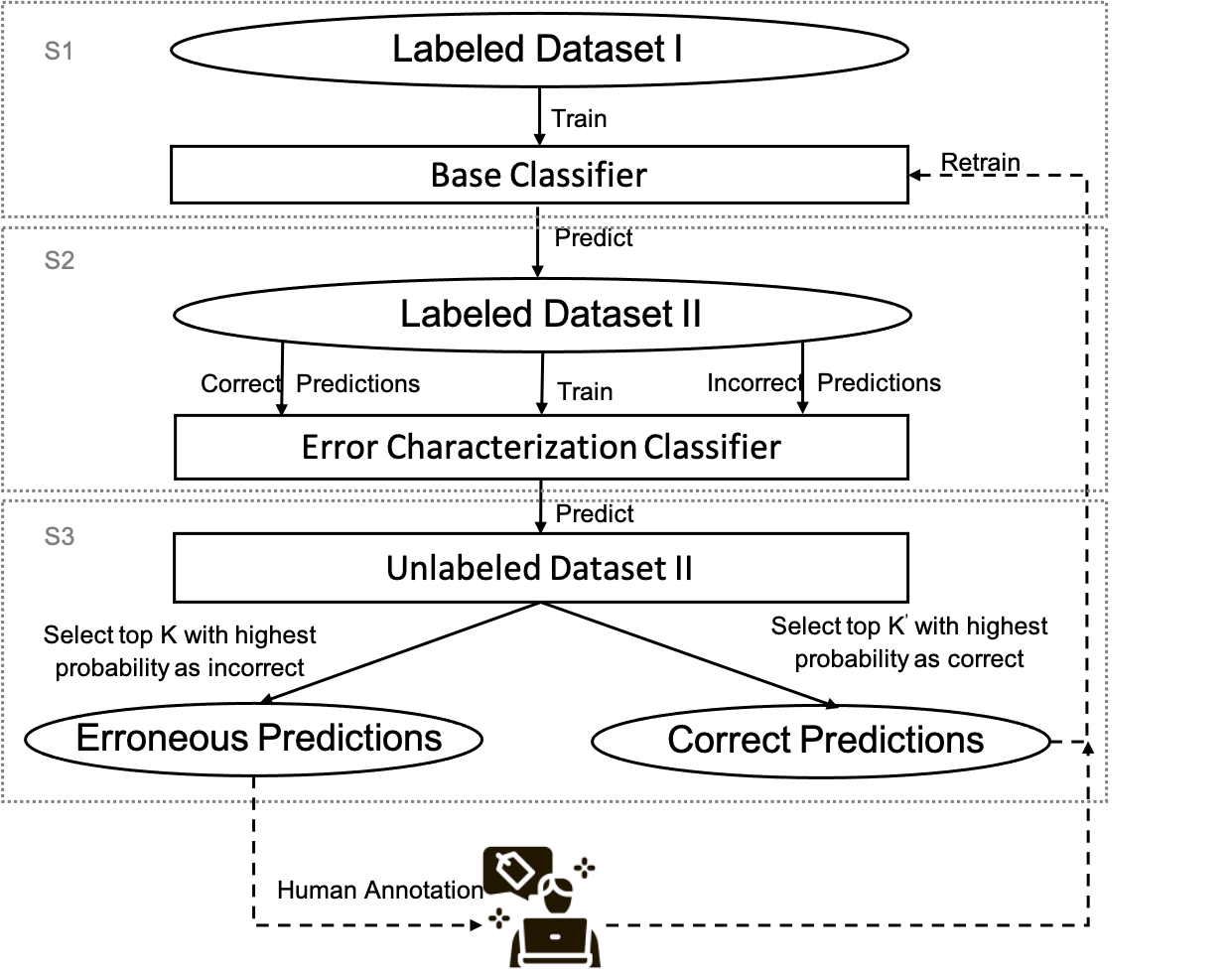}
	\caption{The overall workflow of AEC. Dashed lines represent planned future work}
	\label{workflow}
\end{figure}

Figure~\ref{workflow} summarizes our overall method for constructing a human understandable classifier that can be used to explain the erroneous predictions of a deep  neural  network classifier and thus to improve the model performance. Our method consists of the following steps:\\


\begin{enumerate}
	\item[{\bf S1}:] Train a neural network based  classifier using labeled dataset I, call it as the base classifier.
	\vspace{-.1in}
	\item[{\bf S2}:] Apply the base classifier on another labeled dataset II to get correct and incorrect prediction cases, based on which train a second 2-class error identification classifier with a set of human understandable features. Note here labeled dataset I and II can be in the same domain or in different domains.
	\vspace{-.1in}
	\item[{\bf S3}:] Rank the features according to their individual predictive power.  Apply the error identification classifier from step 2, to a set of unlabeled data from the same domain as labeled dataset II and rank the unlabeled instances according to their prediction probability of being erroneous. These represent the most informative samples that can be further used in an active learning setting.
\end{enumerate}
The focus of the current work is to identify and characterize the error cases  of a base classifier in an human understandable manner. The following two sections describe the experiments and implementation of the framework using a sentiment prediction task as case study. The integration of these samples into an iterative training set up is a work in progress for future extension.

\section{Machine Learning Experiments and Results}
\subsection{Data}

We adopt a cross-domain sentiment analysis task as case study in this section to demonstrate the AEC method, although the proposed method would also be applicable to datasets from the same domain. We chose the cross-domain sentiment analysis task here as it is a challenging, but necessary task within the NLP domain and there are high chances of observing erroneous predictions. We use data from  two different domains, Stanford Sentiment Treebank (SST) \cite{socheretal13} (Labeled Dataset I) to train the base classifier, and a conversational Kaggle Airlines dataset (Labeled + Unlabeled Dataset II) to build and evaluate the error characterization classifier. The conversation domain  represents a new dataset seeking an improvement on the base classifier trained using sentiment reviews.\\
{\bf SST dataset}:  A dataset of movie reviews  annotated at 5 levels  (very  negative,  negative,  neutral,  positive,  and  very  positive).  Sentence level annotations are extracted using the  python package \emph{pytreebank} \footnote{\url{https://pypi.org/project/pytreebank}}. We merged the {\it negative and very-negative}  class labels  into a  single negative class,  {\it positive  and  very-positive}  into a single positive class, keeping neutral as it is.  A  preprocessing  step to remove near duplicates gives a training set distribution as shown in Table~\ref{SST-data}. This is the only dataset used to train the base classifier. \\
\begin{table}
	\begin{small}
		\begin{tabular} { | p{1.50cm} | p{1.2cm} | p{1.2cm} | p{1.2cm} | }
			\hline \bf 	 DataSet & \bf Negative & \bf Neutral & \bf Positive \\ \hline
			  SST & 3304  &  1622& 3605\\
			\hline
	\end{tabular}
	\vspace{-.1in}
\caption{SST dataset distribution }
\label{SST-data}
	\end{small}
\end{table}

{\bf Twitter Airline Dataset}: The dataset is available through the library {\it Crowdflower’s Data for Everyone. \footnote{\url{https://appen.com/resources/datasets/}}} Each tweet is classified as either positive, neutral, or negative. The label distribution for the Twitter Airline is shown in Table~\ref{Airlines-dist-data}. 

\begin{table}
	\begin{small}
		\begin{tabular} { | p{1.5cm} | p{1.1cm} | p{1.1cm} | p{1.1cm} | }
			\hline \bf 	 DataSet & \bf Negative & \bf Neutral & \bf Positive \\ \hline
			Airline & 7366  &  2451& 1847\\
			\hline
	\end{tabular}
	\vspace{-.1in}
\caption{Airline dataset distribution }
\label{Airlines-dist-data}
	\end{small}
\vspace{-4mm}
\end{table}

 \subsection{Train the Base Classifier}
\label{base-class}
 We chose Convolution Neural Network (CNN) as a showcase here, as  the base  sentiment classifier to be trained using the SST dataset. However, the framework can be easily adapted to more advanced state of the art classifiers such as BERT \cite{BERT}. A multi-channel CNN architecture is employed to train as it has been shown to work well  on multiple sentiment datasets including SST \cite{Kim14}. The samples are weighted  to account for class imbalance.

\subsection{Train the Error Characterization  Classifier}
\label{err-class}
We next applied the trained base classifier on the training set of a cross-domain dataset as described in Table \ref{Airlines-dist-data} to get the predictions on a sample of 11664 labeled instances of  Airlines dataset.  Predictions from the base model on this Airlines dataset are further divided into two classes based on the ground truth test labels, correct-prediction and incorrect-prediction.  The base classifier has an overall accuracy of 60.09\% on the Airline dataset as shown in Table~\ref{Airlines-data}. A balanced set is created by undersampling the correct predictions  giving a dataset of total 9310 instances.  We use a 80/20 split for training and testing giving a training set of 7448 and a test set of 1862 instances. This train set serves as the input to train the error characterization classifier  with erroneous or not as labels and different collections of explainable features as independent variables. A random forest classifier using  a 5-fold cross validation was used to train  the error characterization  classifier. \cite{Pedregosaetal11}.

\begin{table}[!htbp]
\begin{small}
\vspace{3mm}
	\begin{tabular} { | p{1.8cm} |p{1.5cm} | p{1.5cm} | p{1.2cm} |  }
	\hline \bf Dataset	 &  \bf Total instances & \bf Correct pred. & \bf InCorrect  pred.\\ \hline

    Airline dataset   & 11664 & 7009&  4655 \\ \hline

	\hline
\end{tabular}
\caption{ Performance of the Base classifier on the Airline dataset  }
\label{Airlines-data}
	\end{small}
\vspace{-4mm}
\end{table}
\subsubsection{Features}
\label{feat}
Our features have been inspired by previous work on  sentiment, disagreement, and conversations.
The feature values are normalized by sentence length.\\
\noindent{\bf Generalized  Dependency.}
Dependency  relations  are obtained using the python package \emph{spacy}  \footnote{\url{https://spacy.io}}. Relations are generalized by replacing the words in each dependency relation by their corresponding POS tag \cite{JoshiRose09,Abbottetal11,MisraEW16}. \\
\noindent{\bf Emotion.} Count of words in each of the 8 emotion classes
from the NRC emotion lexicon (anger, anticipation,
disgust, fear, joy, negative, positive, sadness, surprise, and
trust) available from \cite{mohammad2010emotions}.\\
\noindent{\bf Named Entities.} The count of named entities of each entity type obtained from the python package spacy. \\
\noindent{\bf Conversation.} Lexical indicators indicating greetings, thank, apology, second person reference,
questions starting with do, did, can, could, with who, what, where as described  by \cite{OrabyGMBA17}.\\
\subsection{Predict erroneous predictions from unlabeled data}
Once the error characterization classifier was trained with the set of correctly and incorrectly predicted instances, we then apply it to the 20\% test set of the Twitter Airline data, which consists of a total of 1862 instances as described in
section ~\ref{err-class}.  We selected the top K instances with the highest probability of being incorrectly predicted as the erroneous cases. We hide the actual labels on this test set when selecting the instances. The actual labels will be later used to evaluate the performance of the error characterization classifier.

\section{Evaluation and Results}
In terms of identifying erroneous predictions, in our evaluation, we compare the performance of AEC with uncertainty-based sampling, in which the learner computes a probabilistic output for each sample, and select the  samples that the base classifier is the most uncertain about based on probability scores.


\subsection{ Most informative samples for labeling.}
As we are interested in generating  a ranking  of  incorrect predictions for the base classifier from error characterization classifier, we use {\it precision at top k} as the evaluation metrics in here, which is a commonly used metric  in information retrieval, and defined as
P@K=N/K, where N is the  actual number of  errors samples among  top K predicted. We compare the performance of the error characterization classifier and the uncertainty based sampling  on the test set of 1832 instances as shown in Table~\ref{eval}. It
shows the precision at top K where K varies from 10 to 50.  For the first initial 10 samples, the uncertainty based sampling performs marginally better but as we select more samples (rows 2-5) the proposed approach starts outperforming the baseline.

\begin{table}[h]
	\begin{center}
	\begin{tabular}{ | p{1.4cm} | p{2.0cm} | p{1.8cm} |  }
		\hline \bf TOP K & \bf  uncertainty-based P@K& \bf AEC P@K \\ \hline
		10&0.8&0.7  \\ \hline
		20&0.75&0.8  \\ \hline
		30&0.77&0.83 \\ \hline
		40&0.75&0.83 \\ \hline
		50&0.74&0.76 \\ \hline
	\end{tabular}
	\caption{ Comparison of uncertainty-based sampling (Baseline) with proposed AEC on the test set.}
\label{eval}
	\end{center}
\end{table}

\subsection{Feature Characterization}
When using uncertainity based sampling, it is not always evident why certain samples were selected, or how these samples map to actual errors of the base classifier. In contrast, AEC framework incorporates explainability into sample selection by mapping highly ranked feature sets from the error characterization model with the selected error samples.

Table~\ref{sent-ex} shows a few examples of actual errors from the base classifier that  are also predicted to be errors on the test set from the error characterization classifier.  Words in bold show a few of these  feature mappings. For example, feature set of Row-1 has higher values for the feature {\it question-starters}, text of Row-3 contains  {\it Named Entity type: time,}  a feature present in  highly ranked feature-set of the error characterization classifier as shown in Table~\ref{tevals}.
\begin{table}
\begin{center}
		\begin{scriptsize}
			\begin{tabular}  { | p{0.3cm} | p{3.3cm} | p{0.8cm} | p{0.8cm} | p{0.8cm} |   } \hline
				\bf {S.No}& \bf  {Text} & {Base Pred.} & Actual Label  & Error.  Prob\\ \hline
				1 &@username{\bf  if you could} change your name to @southwestair and {\bf do what }they do...that'd be {\bf awesome}. Also this plane {\bf smells} like onion rings. &Neutral &Negative& 0.84    \\ \hline  
				2& @username now on hold for {\bf 90 minutes }&Neutral &Negative&    0.82    \\ \hline
				3&  @username user is a {\bf compassionate} professional! Despite the flight challenges she made passengers feel like priorities!!& Neutral &Positive &  0.79        \\ \hline
			\end{tabular}
		\end{scriptsize}
\caption{ A subset of  most informative samples for the Base classifier based on error characterization  classifier probability score for the error class.}
\label{sent-ex}
\end{center}
\end{table}

\vspace{-.1in}
\begin{table}[h]
\begin{center}
   \begin{scriptsize}
		\begin{tabular}{| p{1.0cm} | p{4.4cm} |  } \hline
		 {\bf Feature Type} & {\bf Highly ranked features  }\\ \hline
		    Lexical & second\_person, question\_yesno, question\_wh
			!, ?,thanks, no\\ \hline
			NRC& positive, negative, trust, fear, anger, \\ \hline
			Entities & Org, Time , Date, Cardinal\\ \hline
			Dependency & amod-NN-JJ, nummod-NNS,CD, compound-NN-NN, ROOT-NNP-NNP, advmod-VB-RB
		     compound-NN-NNP, neg-VB-RB, amod-NNS,JJ, ROOT-VBN-VBN	    \\  \hline
		\end{tabular}
\end{scriptsize}
\caption { A subset of top 100 Features from  Random Forest.}
\label{tevals}
\end{center}
\end{table}

\section {Conclusion and Future Work}
We present an error characterization framework, called AEC, which allows the model users and developers to understand when and where a model fails. AEC is trained on human understandable linguistic features with erroneous predictions from the base classifier as training input. We used a cross-domain sentiment analysis task as case study to showcase the effectiveness of AEC in terms of error detection and characterization. Our experiments showed that AEC  outperformed uncertainty based sampling in terms of selecting the erroneous samples for continuous model improvements (a strong active learning baseline for selecting the most uncertain samples for continuous model improvements) for the task of predicting errors which can act as most informative samples of the  base classifier. In addition, errors automatically detected by AEC seemed to be more understandable to the model developers. Having these explanations lets the end users make a more informed decision, as well as guide the  labeling decisions for next round of training.  As  our  initial results on sentiment dataset look promising for both performance and explainability, we  are in the process of extending the framework  to run the algorithm iteratively on multiple datasets. While applying the error characterization classifier on the unlabeled datasets, not only we will select the top $K^\prime$ instances with the highest prediction probability of being correctly predicted and add them back to the original training dataset for retraining purpose, but we will also select top $K$ instances with the highest prediction probability of being incorrectly predicted. We will assign those instances to human annotators for labels and add them back to the original labeled data as well for the next iteration of training process. We  will continuously feed these samples to train the base network, and evaluate the actual performance gains for the base classifier.

\bibliographystyle{acl_natbib}
\bibliography{sentiment}
\end{document}